\documentclass{article}

\usepackage[final]{nips_2016}

\usepackage[utf8]{inputenc} 
\usepackage[T1]{fontenc}    
\usepackage{textcomp}
\usepackage{url}            
\usepackage{booktabs}       
\usepackage{amsfonts}       
\usepackage{nicefrac}       
\usepackage[activate={true,nocompatibility},final,kerning=true,spacing=true,factor=1100,stretch=10,shrink=10]{microtype} 
\usepackage{amsmath}
\usepackage{amsthm}
\usepackage{listings}
\usepackage{graphicx}
\usepackage{xspace}
\usepackage{multirow}
\usepackage{array}
\usepackage{bbm}
\usepackage{amsmath}
\usepackage{bm}
\usepackage{adjustbox}
\usepackage{rotating}
\usepackage{afterpage}
\usepackage{pdflscape}

\usepackage[paperVersion]{marcpaper}

\lstnewenvironment{python}[1][]
{
\pythonstyle
\lstset{#1}
}
{}

\DeclareFixedFont{\ttb}{T1}{txtt}{bx}{n}{8} 
\DeclareFixedFont{\ttm}{T1}{txtt}{m}{n}{8}  

\usepackage{color}
\definecolor{deepblue}{rgb}{0,0,0.5}
\definecolor{deepred}{rgb}{0.6,0,0}
\definecolor{deepgreen}{rgb}{0,0.5,0}

\newcommand{\rev}[1]{#1}

\newcommand\pythonstyle{\lstset{
language=Python,
basicstyle=\small\ttm,
otherkeywords={self, with, as},             
keywordstyle=\ttb\color{deepblue},
numbers=left,
stepnumber=1,
firstnumber=1,
numberfirstline=false,
numberstyle=\tiny,
emph={@CompileMe, Param, Var, Input, Output},          
emphstyle=\ttb\color{deepred},    
commentstyle=\color{deepgreen},
frame=tb,                         
showstringspaces=false            %
}}

\usepackage{color}
\newcommand{\langname}{{\sc TerpreT}\xspace} 
\def\arraystretch{1.5}

\usepackage{commands}

\title{\langname: A Probabilistic Programming Language for Program Induction}

\renewcommand{\arraystretch}{1.1}
\author{
\begin{tabular}{c}
Alexander L. Gaunt$^1$, 
Marc Brockschmidt$^1$,
Rishabh Singh$^1$, \\
Nate Kushman$^1$,
Pushmeet Kohli$^1$,
Jonathan Taylor$^2$\thanks{Work done while author was at Microsoft Research.},
Daniel Tarlow$^1$
\end{tabular}\\
$^1$Microsoft Research\hspace{1cm} $^2$perceptiveIO\\
\texttt{\{t-algaun}, \texttt{mabrocks}, \texttt{risin}, \texttt{nkushman}, \texttt{pkohli}, \texttt{dtartlow}\texttt{\}@microsoft.com}\\
\texttt{jtaylor@perceptiveio.com}
}

\begin{document}
(This is a shortened workshop version of the paper at \url{https://arxiv.org/abs/1608.04428}).

\maketitle
\vspace{-0.5cm}
\begin{abstract}
  We study machine learning formulations of inductive program
  synthesis; that is, given input-output examples,
  synthesize source code that maps inputs to corresponding outputs.
  Our key contribution is \langname, a domain-specific
  language for expressing program synthesis problems. A \langname~model
  is composed of a specification of a program representation
  and an interpreter that describes how programs map
  inputs to outputs.
  The inference task is to observe a set of input-output examples and
  infer the underlying program.
  From a
  \langname~model we automatically perform inference using
  four different back-ends:
  gradient descent (thus each \langname~model can be seen as defining a differentiable
  interpreter), linear program (LP) relaxations for graphical
  models, discrete satisfiability solving, and the \sketch~program
  synthesis system.
  \langname has two main benefits. First,
  it enables rapid exploration of a range of domains, program
  representations, and interpreter models.  Second, it separates the
  model specification from the inference algorithm, allowing proper
  comparisons between different approaches to inference.

  We illustrate the value of \langname~by developing several interpreter
  models and performing an extensive
  empirical comparison between alternative inference algorithms
  on a variety of program models.
  To our knowledge, this is the first work to compare gradient-based search
  over program space to traditional search-based alternatives.
  Our key empirical finding is that  constraint solvers dominate the gradient descent and
  LP-based formulations.
\end{abstract}
\vspace{-0.5cm}

\section{Introduction}

Learning computer programs from input-output examples, or Inductive
Program Synthesis (IPS), is a fundamental problem in computer science,
dating back at least to \cite{summers1977methodology} and
\cite{biermann1978inference}.  The field has produced many successes,
with perhaps the most visible example being the FlashFill system in
Microsoft Excel~\citep{gulwani2011automating,cacm12}.

There has also been significant recent interest in neural network-based
models with components that resemble computer programs
\citep{Giles89,Joulin15,grefenstette2015learning,Graves14,Weston14,kaiser2015neural,Reed15,Neelakantan15,Kurach15,andrychowicz2016learning}.
These models combine neural networks with external memory, external computational primitives, and/or built-in structure that reflects a desired algorithmic structure in their execution.
However, none produce programs as output. Instead, the program is hidden inside ``controllers'' composed of neural networks that decide which operations to perform, and the learned program can only be understood in terms of the executions that it produces on specific inputs.

\rev{In this work we focus on models that represent programs as simple, natural source code \citep{hindle2012naturalness}, i.e., the kind of source code that people write.
There are two main advantages to representing programs as source code rather than weights of a neural network controller.
First, source code is interpretable; the resulting models can be inspected by a human and debugged and modified.
Second, programming languages 
are designed to make it easy to express the algorithms that people want to write.
By using these languages as our model representation, we inherit inductive biases that can lead to strong generalization (i.e., generalizing successfully to test data that systematically differs from training).
Of course, natural source code is likely not the best representation in all cases.
For example, programming languages are not designed for writing programs that classify images, and there is no reason to expect that natural source code would impart a favorable inductive bias in this case.}

\rev{Optimization over program space is known to be a difficult problem.
However, recent progress in neural networks showing that it is possible to learn models with differentiable computer architectures, along with the success of gradient descent-based optimization, raises the question of whether gradient descent could be a powerful new technique for searching over program space.}

These issues motivate the main questions in this work.
We ask (a) whether new models can be designed specifically to synthesize interpretable source code that may contain looping and branching structures, and (b) how searching over program space using gradient descent compares to the combinatorial search methods from traditional IPS.

To address the first question we develop models inspired by
intermediate representations used in compilers like LLVM
\citep{lattner2004llvm} that can be trained by gradient descent. These
models interact with external storage, handle non-trivial control flow
with explicit \code{if} statements and loops, and, when appropriately
discretized, a learned model can be expressed as interpretable source
code.  We note two concurrent works, Adaptive Neural
Compilation~\citep{Bunel16} and Differentiable Forth~\citep{Riedel16},
which implement similar models.

To address the second question, concerning the efficacy of gradient
descent, we need a way of specifying many IPS problems such that the
gradient based approach can be compared to alternative
approaches in a like-for-like manner across a variety of domains.
The alternatives
originate from rich histories of IPS in the programming languages
and inference in discrete graphical models. To our
knowledge, no such comparison has previously been performed.

\begin{figure}[t]
\includegraphics[width=\textwidth]{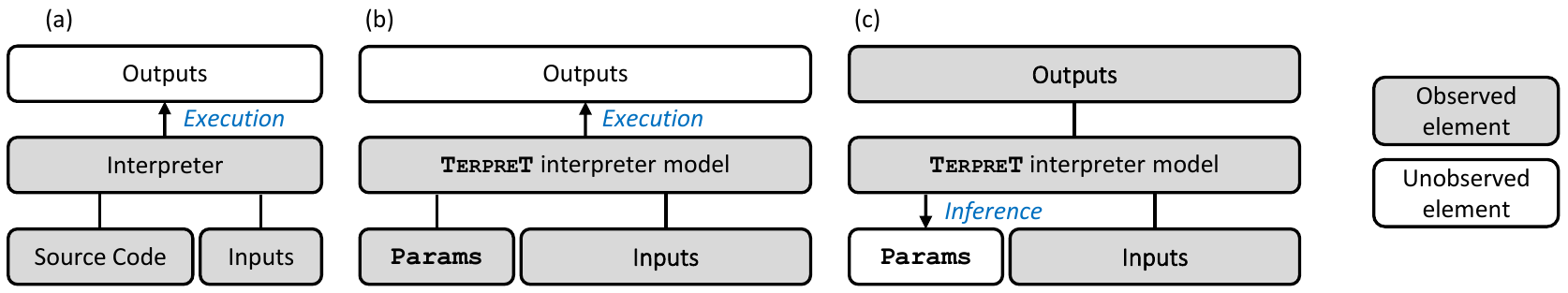}
\caption{A high level view of the program synthesis task. (a) Forward execution of a traditional interpreter.
  (b) Forward execution of a \langname~model. (c) Inference in \langname~model.
  \label{fig:highlevel}
}
\end{figure}

\begin{table}[t]
\hspace{-.15in}
{
  \footnotesize
\begin{tabular}{p{1.2in}lp{3.0in}}
\toprule
Technique name & Optimizer/Solver & Description\\
\midrule
FMGD\newline(Forward marginals, gradient descent) & TensorFlow & A gradient descent based approach which generalizes the approach used by \cite{Kurach15}.\\
(I)LP\newline((Integer) linear programming) & Gurobi & A novel linear program relaxation that supports Gates \citep{minka2009gates}. \\
SMT & Z3 & Translation of the problem into a first-order logical formula with existential constraints.\\
$\sketch$\newline\citep{sketch} & $\sketch$ & Cast the \langname model as a partial program (the interpreter) containing holes (the source code) to be inferred from a specification (input-output examples).\\
\bottomrule
\end{tabular}
\smallskip
\caption{\label{tbl:backends}\langname back-end inference algorithms.}
}
\vspace{-5ex}
\end{table}

Both of the above benefit from being formulated in the context of \langname.
\langname provides a means for describing an \emph{execution model} (e.g., a
Turing Machine, an assembly language, etc.) by defining a
a program representation and an interpreter that
maps inputs to outputs using the parametrized program.  The \langname
description is independent of any particular inference algorithm.
The IPS task is to infer the execution model parameters
(the program) given an execution model and pairs of inputs and
outputs.  An overview of the synthesis task appears in Fig.~\ref{fig:highlevel}.
To perform inference, \langname is automatically compiled into an
intermediate representation which can be fed to a particular inference
algorithm. Table~\ref{tbl:backends} describes the inference algorithms
currently supported by \langname.
Interpretable source code can be obtained directly from the
inferred model parameters.  The
driving design principle for \langname is to strike a subtle balance
between the breadth of expression needed to precisely capture a range
of execution models, and the restriction of expression needed to
ensure that automatic compilation to a range of different back-ends is
tractable.

\section{\langname~Language}
\begin{figure}[t]
\vspace{-1cm}
\begin{tabular}{p{3.3in}p{1.25in}}
{\begin{python}
T = 5
#################################################
#  Source code parametrisation                  #
#################################################
ruleTable = Param(2)[2, 2]

#################################################
#  Interpreter model                            #
#################################################
tape = Var(2)[T]
initial_tape = Input(2)[2]
final_tape = Output(2)
tape[0].set_to(initial_tape[0])
tape[1].set_to(initial_tape[1])

for t in range(1, T - 1):
    with tape[t] as x1:
        with tape[t - 1] as x0:
            tape[t + 1].set_to(ruleTable[x0, x1])

final_tape.set_to(tape[T - 1])
\end{python}} & \raisebox{-1.05\height}{\hspace{0.3cm}\includegraphics[width=1.5in]{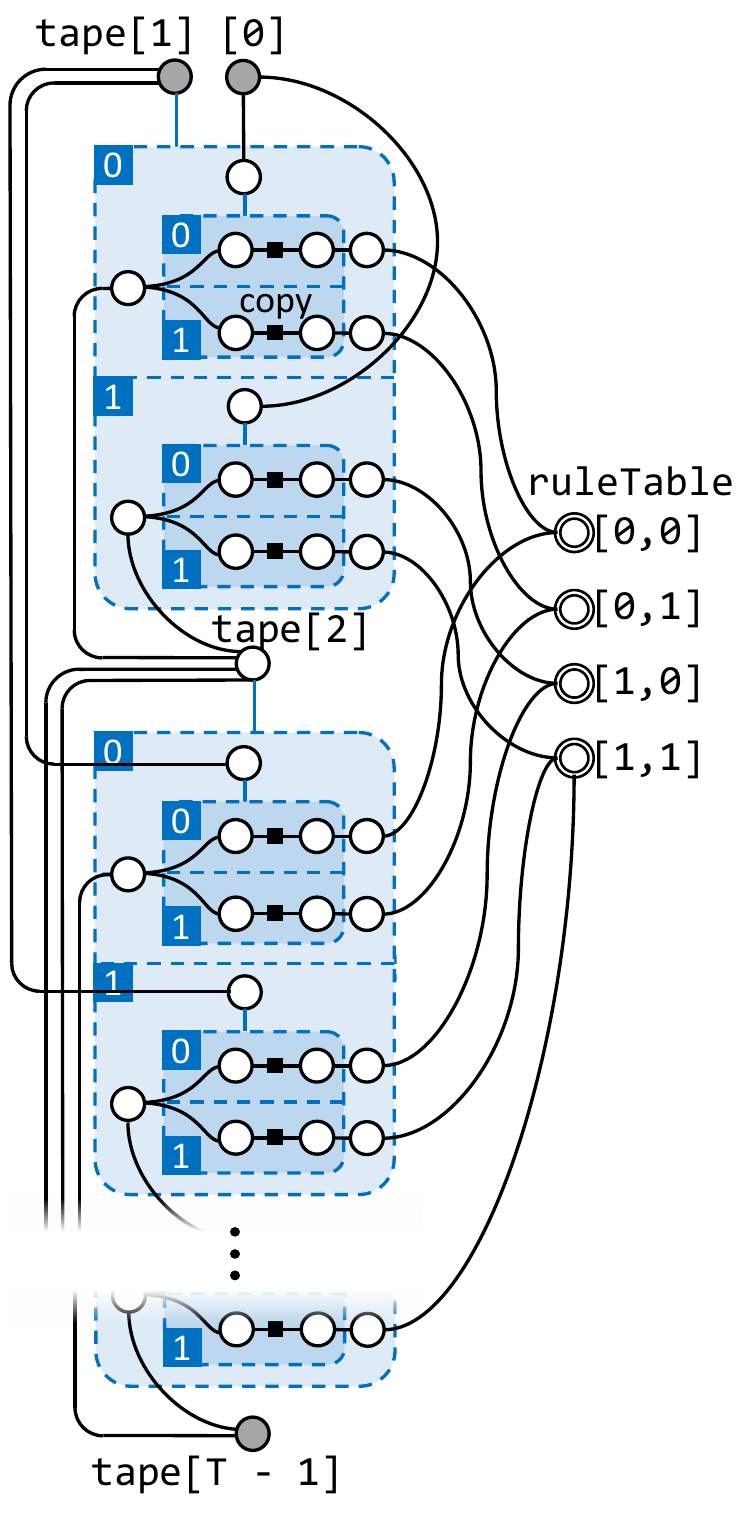}}
\end{tabular}
\vspace{-0.5cm}
\caption{Illustrative example of a \langname~model and corresponding factor graph which describe a toy automaton that updates a binary \texttt{tape} according to the previous two entries and a \texttt{rule} (refer to long version for definition of graphical symbols).
\label{fig:automaton1}
}
\end{figure}

\vspace{-0.2cm}
Due to space, we give just a simple example of how \langname~models are written.
For a full grammar of the language and several longer examples, see the long version \cite{gaunt2016terpret}.
\langname~obeys Python syntax, and we use the Python \texttt{ast} library to parse and
compile \langname~models.
The model is composed of \texttt{Param} variables that define the program, \texttt{Var} variables that
represent the intermediate state of computation, and \texttt{Input}s and \texttt{Output}s. Currently, all
variables must be discrete with constant sizes.
\langname~supports loops with ranges defined by constants\footnote{\rev{These constants define maximum quantities, like the maximum number of timesteps that a program can execute for, or the maximum number of instructions that can be in a program.
Variable-length behavior can be achieved by defining an absorbing end-state, by allowing a no-op instruction, etc.}},
    if statements, arrays, array indexing,
and user-defined functions that map a discrete set of inputs to a discrete output.
In total, the constraints imply that a model can be converted into a gated factor graph \citep{minka2009gates}.
The details of how this works are in the longer version.
A very simple \langname~model appears in Fig.~\ref{fig:automaton1}. In this model, there is a binary
tape that the program writes to.
The program is a rule table that specifies which state to write at the current position of
a tape, conditional on the pattern that appeared on the two previous tape locations.
The first two tape positions are initialized as Inputs, and the final tape position is the program output.

We have used \langname~to build much more complicated models, including a Turing Machine, boolean circuits, a Basic Block model that is similar to the intermediate representation used by LLVM, and an assembly-like model.
\vspace{-0.2cm}

\section{Experimental Results}
\vspace{-0.2cm}
Table~\ref{tbl:benchmarkTasks} gives an overview of the benchmark programs which we attempt to synthesize. We created \langname descriptions for each execution model, we designed three synthesis tasks per model which are specified by up to 16 input-output examples. We measure the time taken by the inference techniques listed in Table~\ref{tbl:backends} to synthesize a program for each task. Results are summarized in Table~\ref{tbl:benchmarkResults}.

\begin{table}[t]
\vspace{-0.8cm}
\hspace{-2.5cm}
\begin{tabular}{c}
\renewcommand{\arraystretch}{1.3}
\begin{tabular}{p{1.8in}c}
\centering
\raisebox{-0.5\height}{\includegraphics[width=1.7in]{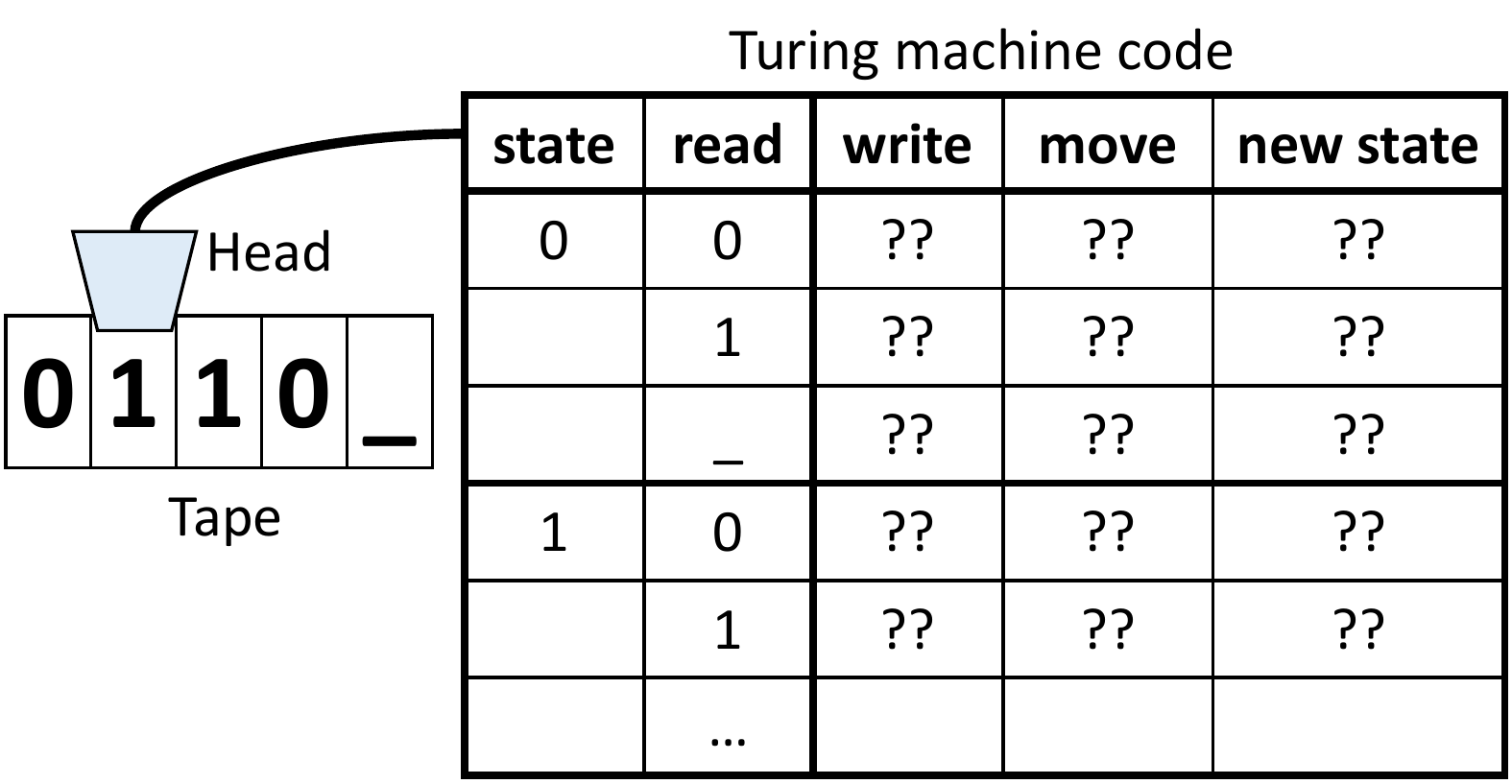}} &
\begin{tabular}{p{1.57in}p{3.1in}}
\toprule
TURING MACHINE & Description\\
\midrule
Invert & Perfom bitwise inversion of a binary string on the tape.\\
Prepend zero & Right shift all symbols and insert a ``\code{0}" in the first cell.\\
Binary decrement &  Decrement a binary representation on the tape by 1.\\
\bottomrule
\end{tabular}
\end{tabular}\\
\renewcommand{\arraystretch}{1.3}
\begin{tabular}{p{1.8in}c}
\centering
\raisebox{-0.5\height}{\includegraphics[width=1.7in]{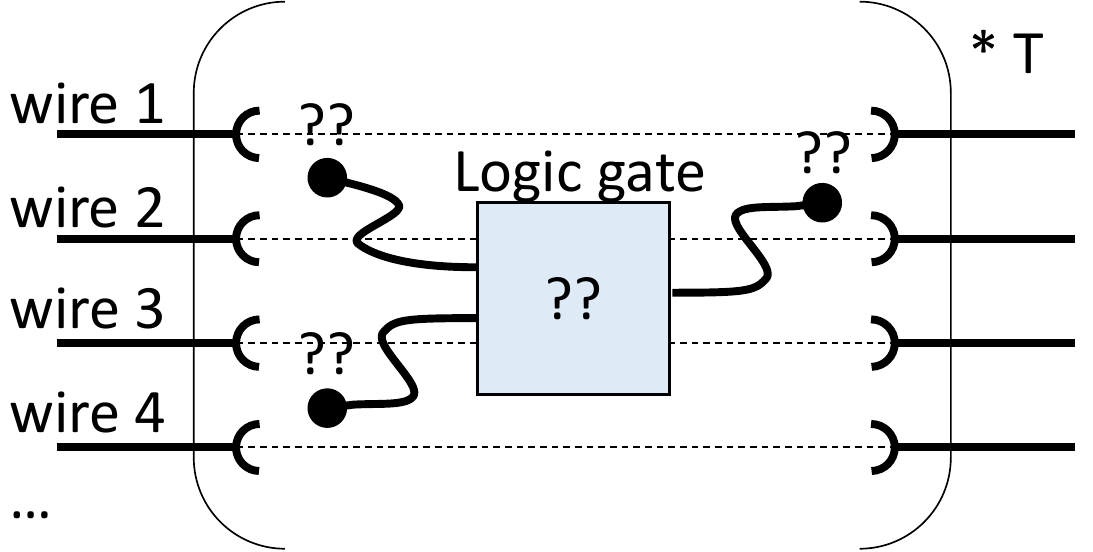}} &
\begin{tabular}{p{1.57in}p{3.1in}}
BOOLEAN CIRCUITS& Description\\
\midrule
2-bit controlled shift register & Swap bits on wires 2 \& 3 iff wire 1 is on.\\
full adder &  Perform binary addition of two bits including carries.\\
2-bit adder &   Perform binary addition on two-bit numbers.\\
\bottomrule
\end{tabular}
\end{tabular}\\
\renewcommand{\arraystretch}{1.3}
\begin{tabular}{p{1.8in}c}
\centering
\raisebox{-0.5\height}{\includegraphics[width=1.5in]{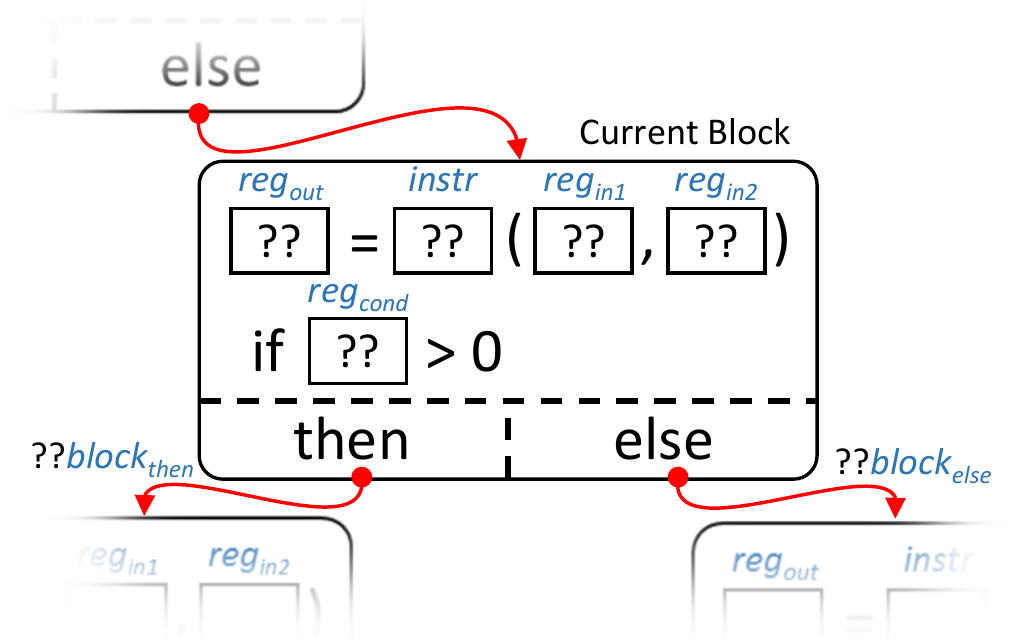}} &
\begin{tabular}{p{1.57in}p{3.1in}}
BASIC BLOCK &  Description\\
\midrule
Access &  Access the $k^{\rm th}$ element of a contiguous array.\\
Decrement &  Decrement all elements in a contiguous array.\\
List-K &  Access the $k^{\rm th}$ element of a linked list.\\
\bottomrule
\end{tabular}
\end{tabular}\\
\renewcommand{\arraystretch}{1.3}
\begin{tabular}{p{1.8in}c}
\centering
\raisebox{-0.5\height}{\includegraphics[width=1.6in]{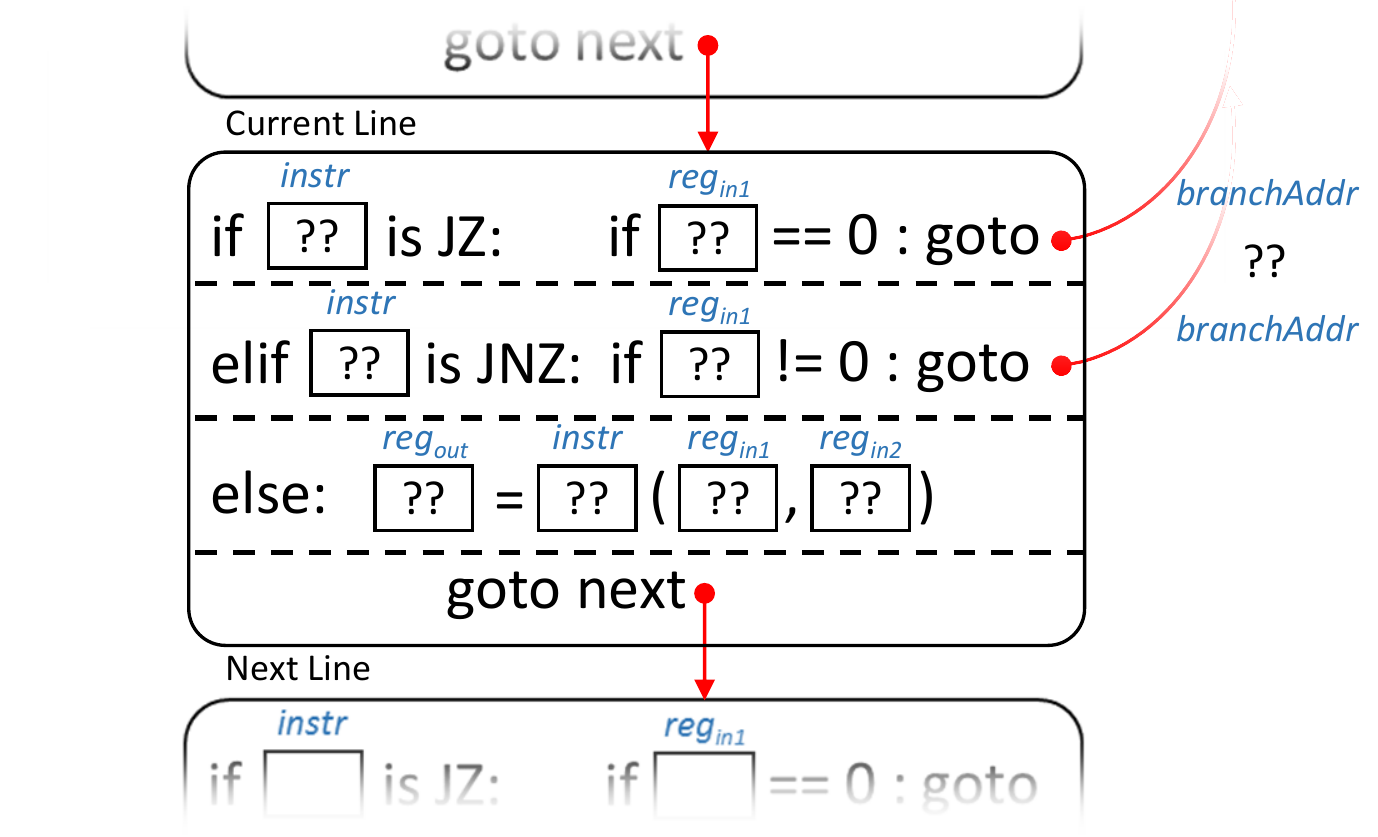}} &
\begin{tabular}{p{1.57in}p{3.1in}}
ASSEMBLY & Description\\
\midrule
Access & \\
Decrement & \hspace{1cm}As above.\\
List-K &\\
\bottomrule
\end{tabular}
\end{tabular}
\end{tabular}
\caption{\label{tbl:benchmarkTasks} Overview of benchmark problems, grouped by execution model. We illustrate the basic structure of each model, with parameters of the model to be inferred denoted by ??. }
\vspace{-0.6cm}
\end{table}

With the exception of the FMGD algorithm, we perform a single run with a timeout of 4 hours for each
task. For the FMGD algorithm we run both a Vanilla form and an Optimized form with additional heuristics such as gradient clipping, gradient noise and an entropy bonus (\cite{Kurach15}) to aid convergence. Even with these heuristics we observe that several random initializations of the FMGD algorithm stall in an uninterpretable local optimum rather than finding an interpretable discrete program in a global optimum. In the Vanilla case we report the fraction of 20 different random initializations which
lead to \rev{any globally optimal solution consistent with the input-output specification} and also the wall clock time for 1000 epochs of the gradient descent
algorithm (the typical number required to reach convergence on a successful run).
In the Optimized FMGD case, we randomly draw 120 sets of hyperparameters from a manually chosen distribution and run the learning
with 20 different random initializations for each setting. We then report the success rate for the best hyperparameters found and also for the average across all
runs in the random search.

Our results show that traditional techniques employing constraint solvers
(SMT and Sketch) outperform other methods (FMGD and ILP), with $\sketch$ being the only system
able to solve all of these benchmarks before timeout\footnote{\rev{Additional (pre)processing performed by \sketch~makes it slower than a raw SMT solver on the smaller tasks but allows it to succeed on the larger tasks}}. 
Furthermore, Table~\ref{tbl:benchmarkResults} highlights that the precise formulation of the interpreter model can
affect the speed of synthesis. Both the Basic Block and Assembly models are equally expressive, but
the Assembly model is biased towards producing straight line code with minimal branching. In cases
where synthesis was successful the Assembly representation is seen to outperform the Basic Block model
in terms of synthesis time.

In addition, we performed a separate experiment to investigate the local optima arising in the FMGD formulation. Using \langname we describe the task of inferring the values of a bit string of length $K$ from observation of the parity of neighboring variables. 
It is possible to show analytically that the number of local optima grow exponentially with $K$, and Table \ref{tbl:parity-chain} provides empirical evidence that these minima are encountered in practice and they
hinder the convergence of the FMGD algorithm. 
\vspace{-0.3cm}

\section{Conclusion}
\vspace{-0.3cm}
We presented \langname, a probabilistic programming language for specifying IPS problems.
The flexibility of the \langname language in combination with the four inference backends allows like-for-like comparison between gradient-based and constraint-based techniques for inductive program synthesis. The primary take-away from the experiments is that constraint solvers outperform other approaches in all cases studied. However, our work is (intentionally) only measuring the ability to efficiently search over program space. \rev{We remain optimistic about extensions to the \langname framework which allow differentiable interpreters to handle problems involving perceptual data \citep{gaunt16}, and about using machine learning techniques to guide search-based techniques \citep{balog16}}.

\afterpage{
\begin{landscape}
\begin{table}
\begin{center}
\renewcommand{\arraystretch}{1.1}
\begin{tabular}{lccc|cccc|ccc}
&&&&\multicolumn{4}{c|}{FMGD}& ILP & SMT & $\sketch$\\
& $\log_{10}(D)$ & $T$ & $N$ & Time & Vanilla & Best Hypers & Average Hypers & Time & Time & Time \\
\toprule
TURING MACHINE\\
\midrule
Invert & 4 & 6 & 5 & 76.5 & 100\% & 100\% & 51\% & 0.6 & 0.7 & 3.1 \\
Prepend zero & 9 & 6 & 5 & 98 & 60\% & 100\% & 37\% &  17.0 & 0.9 & 2.6\\
Binary decrement & 9 & 9 & 5 &  163 & 5\% & 25\% & 2\% & 191.9 & 1.6 & 3.3\\
\bottomrule
BOOLEAN CIRCUITS\\
\midrule
2-bit controlled shift register & 10 & 4 & 8 & - & - & - & - & 2.5 & 0.7 & 2.7\\
Full adder & 13 & 5 & 8 & - & - & - & - &  38 & 1.9 & 3.5\\
2-bit adder & 22 & 8 & 16 & - & - & - & - &  13076.5 & 174.4 & 355.4\\
\bottomrule
BASIC BLOCK\\
\midrule
Access & 14 & 5 & 5 & 173.8 & 15\% & 50\% & 1.1\%& 98.0 & 14.4 & 4.3\\
Decrement & 19 & 18 & 5 &  811.7  & - & 5\% & 0.04\% & - & - & 559.1\\
List-K & 33 & 11 & 5 & - & - & - & - & - & - & 5493.0\\
\bottomrule
ASSEMBLY\\
\midrule
Access & 13 & 5 & 5 & 134.6 & 20\% & 90\% & 16\% & 3.6 &  10.5 & 3.8\\
Decrement & 20 & 27 &  5 & - & - & - & - & - & - & 69.4\\
List-K & 29 & 16 & 5 & - & - & - & - & - & - & 16.8\\
\bottomrule
\end{tabular}
\end{center}
\caption{\label{tbl:benchmarkResults} Benchmark results. For FMGD we present the time in seconds for 1000 epochs and the success rate out of \{20, 20, 2400\} random restarts in the \{Vanilla, Best Hypers and Average Hypers\} columns respectively. For other back-ends we present the time in seconds to produce a synthesized program. The symbol - indicates timeout ($>4$h) or failure of any random restart to converge.
$N$ is the number of provided input-output examples used to specify the task in each case.
}
\end{table}
\begin{table}
  \begin{center}
\renewcommand{\arraystretch}{1}
  \begin{tabular}{rcccccc}
    \toprule
     & $K=4$  & $K=8$  & $K=16$  & $K=32$  & $K=64$   & $K=128$ \\
    \midrule
    {Vanilla FMGD}   &  100\% & 53\%   & 14\%   & 0\%   & 0\%   & 0\% \\
    {Best Hypers}    &  100\% & 100\%  & 100\%  & 70\%  & 80\%  & 0\% \\
    {Average Hypers}    &  84\% & 42\%  & 21\%  & 4\%  & 1\%  & 0\% \\
    \bottomrule
  \end{tabular}
  \end{center}
  \caption{\label{tbl:parity-chain}
    Percentage of runs that converge to the global optimum for FMGD on
    the Parity Chain example of length $K$.
  }
\end{table}

\end{landscape}
}

\bibliography{references}
\bibliographystyle{plainnat}

\end{document}